\documentclass[sigconf,nonacm]{acmart}
\usepackage[table]{xcolor}
\usepackage{pifont}
\usepackage{subcaption}
\newcommand{\cmark}{\ding{51}}
\newcommand{\xmark}{\ding{55}}
\newcommand{\safeimg}[2]{%
  \IfFileExists{#2}{\includegraphics[width=#1]{#2}}{\fbox{\parbox[c][0.22\textheight][c]{#1}{\centering Missing image: \\ \texttt{#2}}}}%
}
\renewcommand\footnotetextcopyrightpermission[1]{}
\AtBeginDocument{%
  }
\begin{document}

\title{Geo-EVS: Geometry-Conditioned Extrapolative View Synthesis for Autonomous Driving}

\author{Yatong Lan}
\affiliation{%
  \institution{School of Vehicle and Mobility, Tsinghua University}
  \city{Beijing}
  \postcode{100084}
  \country{China}
}
\affiliation{%
  \institution{State Key Laboratory of Intelligent Green Vehicle and Mobility, Tsinghua University}
  \city{Beijing}
  \postcode{100084}
  \country{China}
}
\affiliation{%
  \institution{School of Science, Minzu University}
  \city{Beijing}
  \postcode{100081}
  \country{China}
}

\author{Rongkui Tang}
\affiliation{%
  \institution{Chongqing Changan Automobile Co., Ltd.}
  \city{Chongqing}
  \country{China}
}

\author{Lei He}
\authornote{Corresponding author: Lei He (helei2023@tsinghua.edu.cn).}
\affiliation{%
  \institution{School of Vehicle and Mobility, Tsinghua University}
  \city{Beijing}
  \postcode{100084}
  \country{China}
}
\affiliation{%
  \institution{State Key Laboratory of Intelligent Green Vehicle and Mobility, Tsinghua University}
  \city{Beijing}
  \postcode{100084}
  \country{China}
}


\begin{abstract}
Extrapolative novel view synthesis can reduce camera-rig dependency in autonomous driving by generating standardized virtual views from heterogeneous sensors. Existing methods degrade outside recorded trajectories because extrapolated poses provide weak geometric support and no dense target-view supervision. The key is to explicitly expose the model to out-of-trajectory condition defects during training. We propose \textbf{Geo-EVS}, a geometry-conditioned framework under sparse supervision. Geo-EVS has two components. \emph{Geometry-Aware Reprojection (GAR)} uses fine-tuned VGGT to reconstruct colored point clouds and reproject them to observed and virtual target poses, producing geometric condition maps. This design unifies the reprojection path between training and inference. \emph{Artifact-Guided Latent Diffusion (AGLD)} injects reprojection-derived artifact masks during training so the model learns to recover structure under missing support. For evaluation, we use a \emph{LiDAR-Projected Sparse-Reference (LPSR)} protocol when dense extrapolated-view ground truth is unavailable. On Waymo, Geo-EVS improves sparse-view synthesis quality and geometric accuracy, especially in high-angle and low-coverage settings. It also improves downstream 3D detection.

\end{abstract}

\maketitle

\section{Introduction}
As autonomous driving systems scale, cross-platform data reuse remains a major bottleneck. A primary reason is sensor heterogeneity. Different vehicle lines adopt different camera counts, mounting positions, and fields of view, making data and models tightly coupled to specific rig configurations~\cite{waymoopen2020,caesar2020nuscenes}. As a consequence, launching a new platform often requires expensive recollection and relabeling. This increases both engineering cost and development cycle.

Extrapolative view synthesis offers a direct route to this problem by generating standardized virtual views beyond native camera layouts. If reliable extrapolated views can be produced from existing onboard images, training data can be reorganized into a shared camera space across heterogeneous vehicles. However, this setting is substantially more challenging than in-manifold novel-view synthesis within overlapping view manifolds. In real driving logs, capture is constrained by a single trajectory and yields sparse, biased 3D observations. Once the target viewpoint departs from the observed manifold, dense target-view supervision is unavailable and geometric reprojection becomes severely under-supported.

Existing methods remain unstable in this regime. Diffusion-based approaches learn strong image priors, but may hallucinate textures over geometrically unsupported regions when viewpoint coverage is sparse. Reconstruction pipelines based on NeRF or Gaussian Splatting are geometrically grounded, yet remain sensitive to wide-baseline support sparsity and often exhibit topology breaks, stretched structures, and large invalid regions in extrapolated views. Across both families, a shared limitation persists: extrapolated reprojections contain large holes and discontinuities that are rare in observed-pose training conditions, causing a systematic train--test mismatch.

To address this gap, we propose \textbf{Geo-EVS}, a geometry-conditioned extrapolative synthesis framework under sparse supervision. Geo-EVS combines geometry-consistent construction and artifact-aware generation, as detailed below.

A \emph{Geometry-Aware Reprojection (GAR)} module first builds training pairs in which the condition is a reprojected point-cloud rendering and the supervision label is the target-view supervision-domain image at the same camera pose. Both training and inference therefore use the same geometric reprojection pathway. An \emph{Artifact-Guided Latent Diffusion (AGLD)} backbone performs geometry-to-image synthesis with early spatial alignment to preserve structure while recovering details. During optimization, precomputed reprojection-derived artifact masks are injected into conditions. This trains the model to recover valid structure under missing-support patterns. Together, these designs target reliable extrapolated-view generation as a practical interface for cross-rig data reuse.

Since dense target-view ground truth is unavailable, we establish a \emph{LiDAR-Projected Sparse-Reference (LPSR)} protocol for principled quantitative evaluation. Metrics are computed only on valid projected regions, enabling fair comparison under realistic extrapolation constraints. On the Waymo Open Dataset, Geo-EVS consistently improves sparse-view fidelity and geometric consistency. The generated views further provide measurable gains in downstream 3D detection.

\noindent\textbf{Contributions.}
\begin{itemize}

\item A unified task formulation of autonomous-driving extrapolative view synthesis for cross-rig reuse, where generation is cast as geometry-conditioned learning under sparse supervision rather than dense target-view reconstruction.

\item A GAR-based geometry-aligned data construction strategy that preserves the projection pathway and artifact formation mechanism, enabling tractable training when dense supervision at extrapolated poses is unavailable.

\item An extrapolation-oriented generator, built on AGLD with artifact-aware optimization, that better preserves geometric structure while recovering photorealistic details under large pose gaps and limited support.

\item A practical evaluation protocol (LPSR) for no-dense-ground-truth settings, enabling principled comparison under realistic extrapolation constraints.

\end{itemize}

\section{Related Work}
\paragraph{Novel View Synthesis for Autonomous Driving.}
Autonomous-driving view synthesis has evolved from image-based rendering and layered plane representations~\cite{zhou2018stereo,flynn2016deepstereo} to neural scene models. NeRF-style methods and large-scale extensions (e.g., Mip-NeRF 360, Block-NeRF, Mega-NeRF, CityNeRF) substantially improve realism and scalability in outdoor environments~\cite{mildenhall2020nerf,barron2022mipnerf360,tancik2022blocknerf,turken2022meganerf,zhang2024citynerf}. For dynamic street scenes, motion-aware formulations and Gaussian-based rendering further improve efficiency and temporal modeling~\cite{li2021nsff,teed2021raft3d,suds2023,emernerf2023,drivinggaussian2024,streetgaussians2024}.

These lines of work establish strong reconstruction quality, especially when viewpoint overlap is adequate~\cite{wang2021ibrnet,barron2021mipnerf}. Their common evaluation setting remains predominantly dense-ground-truth or near-trajectory novel-view reconstruction. By comparison, our target setting is extrapolative synthesis with sparse, partially unverifiable supervision regions, which defines a different task objective and evaluation regime rather than a minor variant of standard reconstruction~\cite{kerbl20233dgs,philion2020lss}.

\paragraph{Conditional Generative Modeling for View Synthesis.}
Diffusion models (DDPM, latent diffusion) provide a strong foundation for high-fidelity generation~\cite{ho2020ddpm,rombach2022ldm}. Conditioning strategies such as ControlNet, T2I-Adapter, and GLIGEN show that structural guidance can effectively control generation~\cite{zhang2023controlnet,mou2023t2iadapter,li2023gligen}. In novel-view generation, Zero123, SyncDreamer, and MVDiffusion demonstrate promising viewpoint generalization from sparse observations~\cite{liu2023zero123,syncdreamer2024,tang2023mvdiffusion}. In driving, related generative systems extend conditioning to BEV-, box-, and LiDAR-guided synthesis~\cite{drivedreamer2023,gaia12023,bevgen2024,magicdrive2024}. Our experimental baseline FreeVS also belongs to this diffusion-based family~\cite{freevs2024}.

Why these methods remain insufficient in our setting is more specific than ``diffusion vs. reconstruction''. Existing pipelines often assume cleaner condition distributions than those induced by out-of-trajectory reprojection, so robustness to structured holes/discontinuities is under-modeled~\cite{song2021sde,nichol2021improvedddpm,pathak2016context}. A concise closest-method contrast is as follows. Reconstruction-heavy baselines, including 3DGS~\cite{kerbl20233dgs}, Street Gaussians~\cite{streetgaussians2024}, and EmerNeRF~\cite{emernerf2023}, rely more on overlap-consistent geometry and degrade under sparse support. The diffusion baseline FreeVS~\cite{freevs2024} improves perceptual plausibility but does not explicitly target extrapolation-induced condition defects during training. We therefore position Geo-EVS at the task level: extrapolative synthesis under sparse and partially unverifiable supervision, rather than dense/near-trajectory reconstruction with stable support.

\section{Method}
\subsection{Overview}
\begin{figure*}[t]
  \centering
  \safeimg{0.98\textwidth}{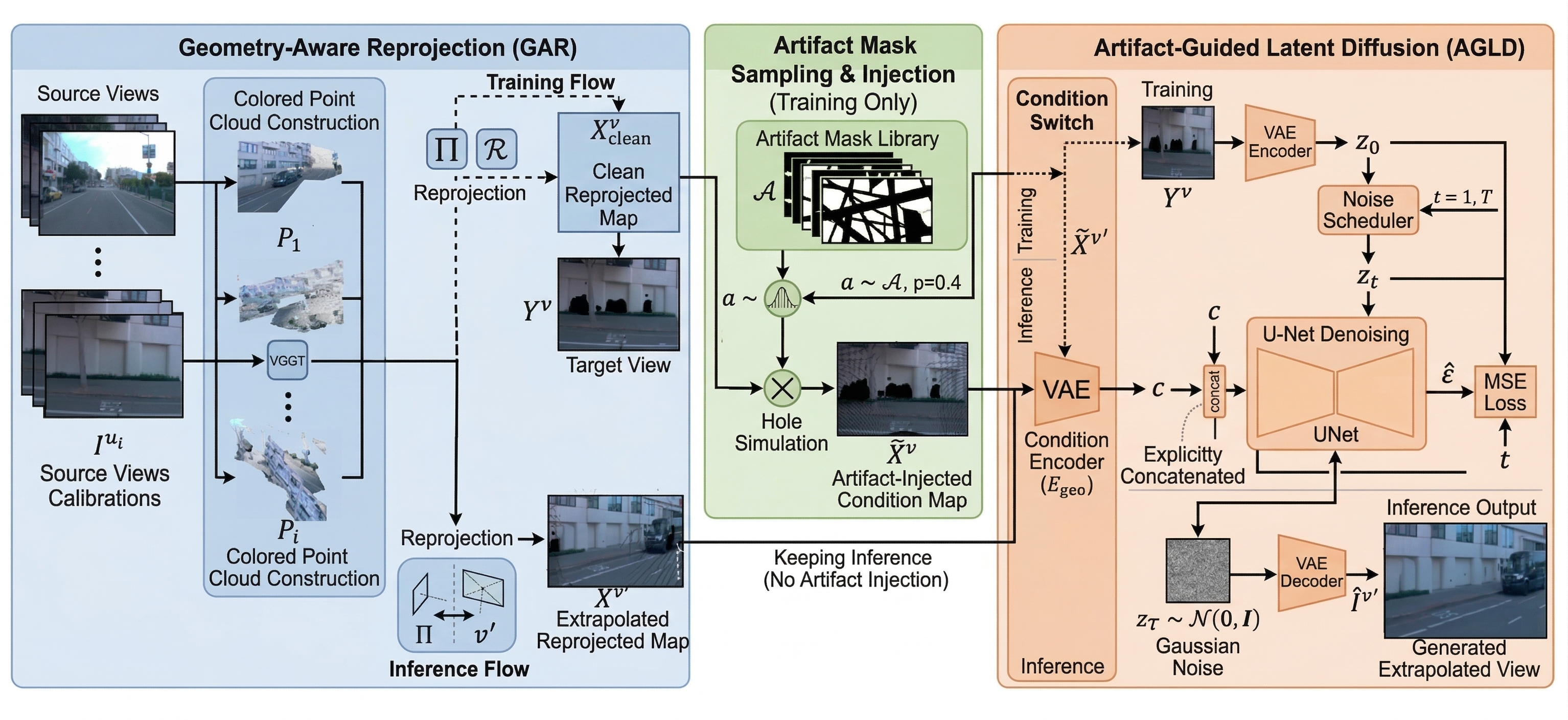}
  \caption{Overview of Geo-EVS. GAR constructs geometry-conditioned maps through reprojection with a unified projection interface for training and inference, and AGLD performs geometry-to-image generation with artifact-aware optimization.}
  \Description{A block diagram showing the pipeline of Geo-EVS, including GAR and AGLD components and their interactions.}
  \label{fig:pipeline}
\end{figure*}

Given a target camera pose $v'$, our goal is to synthesize its target-view image from geometric condition maps derived from observed views. We formulate the task as geometry-conditioned generation:
\begin{equation}
\hat{I}^{v'} = G_\theta(X^{v'}),
\end{equation}
where $X^{v'}\in\mathbb{R}^{H\times W\times 3}$ is a reprojection-based geometric condition map and $G_\theta$ is a conditional latent diffusion model.

Geo-EVS contains three components: (i) geometry-conditioned pair construction, (ii) geometry-to-image diffusion generation, and (iii) artifact-aware robustness training. Training uses observed-pose supervision pairs $(X^v,Y^v)$, while inference applies the same generator architecture to out-of-trajectory geometric condition maps $X^{v'}$. As shown in Fig.~\ref{fig:pipeline}, Sec.~3.2 defines geometric condition map construction, Sec.~3.3 defines conditional diffusion training and inference, and Sec.~3.4 defines artifact-mask injection for robustness.

Our core insight is to recast extrapolative view synthesis from cross-view generation into structure recovery under sparse geometric condition maps. GAR keeps the \emph{projection operator} consistent between training and inference, but the \emph{input distribution} can still shift: extrapolated poses yield geometric condition maps with more missing support than observed poses, inducing a shift between $p(X\mid v\in\mathcal{V}_{\text{obs}})$ and $p(X\mid v\in\mathcal{V}_{\text{ext}})$.

\subsection{Geometry-Conditioned Pair Construction}
\paragraph{Motivation.}
To ensure trainability under no extrapolated-view labels, we first construct condition--label pairs from observed poses only.

\paragraph{Design.}
For a source image $I^u\in\mathbb{R}^{H\times W\times 3}$ with calibration $(K_u,T_u)$, we use \emph{VGGT} (Visual Geometry Grounded Transformer, \cite{vggt2024}), a geometric foundation model for dense 3D-aware geometric prediction, as a geometric feature extractor to provide geometry-consistent point-map priors for GAR. Specifically, we first fine-tune VGGT with fixed camera poses to adapt it to driving-scene geometry, and then freeze it during Geo-EVS training to extract geometric features only (no VGGT parameter update in the diffusion stage). We then estimate a dense point map $\mathbf{M}^u$ and convert it to a colored point cloud. The source for this construction is single-frame (no multi-frame point-cloud fusion is used in GAR):
\begin{equation}
\mathbf{P}=\mathcal{F}(\mathbf{M}^u,I^u).
\end{equation}
In training, we set $u=v$ and project $\mathbf{P}$ back to the same camera pose to obtain a sparse geometric condition map aligned with supervision-domain targets. Cross-view construction at $u\neq v$ would require target-view supervision at those poses, which is unavailable for out-of-trajectory viewpoints; this motivates learning from clean observed-pose pairs and explicitly simulating missing-support patterns in Sec.~3.4. Functionally, GAR is not restricted to interpolation; it defines a unified reprojection interface that can instantiate any virtual target pose, including outward lateral offsets used in our extrapolative experiments. Given a camera $(K_v,T_v)$, we project and rasterize $\mathbf{P}$ as
\begin{equation}
X^v=\mathcal{R}(\Pi(\mathbf{P},K_v,T_v)).
\end{equation}
Here $\Pi(\cdot)$ denotes perspective projection and $\mathcal{R}(\cdot)$ denotes z-buffer rasterization. For training, each pair is formed as $(X^v,Y^v)$, where $Y^v$ is the dense RGB image captured at the same camera pose as $X^v$. At inference, we apply the same generator to $X^{v'}$ rendered at novel poses $v'$.

Projection validity is determined by a deterministic visibility rule. A projected point is considered valid only if it satisfies three constraints: positive depth, in-frame coordinates, and front-most visibility under z-buffering. When multiple 3D points map to the same pixel, we keep the point with minimum depth and discard the others. Pixels without surviving projections are zero-filled in $X^v$, yielding a unique sparse condition map for downstream generation.

This calibration-driven construction is scalable on raw driving logs and provides supervised geometry-to-image pairs without requiring dense labels at extrapolated poses. It also keeps the geometric interface consistent between training and inference. The outputs of this stage, $(X^v, Y^v)$, are directly consumed by later modules. Section~3.3 uses them for diffusion training, and Section~3.4 replaces $X^v$ with $\tilde{X}^v$ while keeping the same training objective.

\subsection{Geometry-to-Image Conditional Diffusion}
\paragraph{Motivation.}
Given the trainable pairs from Sec.~3.2, the next problem is generability: geometric condition maps are sparse and discontinuous, so the generator must preserve structure on valid support while recovering plausible textures in missing regions.

\paragraph{Design.}
We train a latent diffusion model with standard noise-prediction training. For each sample, the supervision target $Y^v$ is encoded to clean latent $z_0$; a timestep $t$ is sampled and Gaussian noise is added to obtain $z_t$. The geometric condition map (clean $X^v$ or artifact-injected $\tilde{X}^v$) is encoded by $\mathcal{E}_{geo}(\cdot)$ to condition features $c$, which are concatenated with $z_t$ at the UNet input:
\begin{equation}
\tilde{z}_t=[z_t\oplus c],
\end{equation}
and optimize the denoiser with the standard objective:
\begin{equation}
\mathcal{L}_{\text{diff}}=\mathbb{E}_{z_0,\epsilon,t}\left[\left\|\epsilon-\epsilon_\theta([z_t\oplus c],t)\right\|_2^2\right].
\end{equation}
In implementation, the training objective uses per-sample weighting consistent with the dataloader:
\begin{equation}
\mathcal{L}_{\text{train}}=\frac{1}{B}\sum_{i=1}^{B} w_i\left\|\epsilon_i-\epsilon_\theta([z_{t,i}\oplus c_i],t_i)\right\|_2^2.
\end{equation}
To enable classifier-free guidance at inference, we apply conditional dropout during training: with probability $p_{\text{drop}}$, the geometric condition is replaced by a null condition token $\varnothing$. Training samples one timestep per instance (standard diffusion training) and does not require unrolling the full denoising chain. Multi-step denoising is used only at inference. At inference, we start from Gaussian noise and denoise for $T$ steps conditioned on $X^{v'}$ to obtain $\hat{I}^{v'}$ after VAE decoding. For completeness, classifier-free guidance can be written as
\begin{equation}
\hat{\epsilon}_{\text{cfg}}(z_t,t,c)=\hat{\epsilon}_{\theta}(z_t,t,\varnothing)+s_{\text{cfg}}\Big(\hat{\epsilon}_{\theta}(z_t,t,c)-\hat{\epsilon}_{\theta}(z_t,t,\varnothing)\Big),
\end{equation}
where $c=\mathcal{E}_{geo}(X^{v'})$ and $s_{\text{cfg}}$ is the guidance scale. In our setting, this reduces to standard conditional denoising when $s_{\text{cfg}}=1$. The noise-prediction objective implicitly fits the conditional distribution $p(Y\mid X)$, so the model learns a structured image prior consistent with the observed geometry. When the condition map contains missing pixels, these locations contribute no geometric constraint; the denoiser therefore relies on the learned prior to complete plausible content while remaining anchored by valid-condition regions. In other words, hole filling is not introduced by an extra inpainting loss but emerges from conditional likelihood training once the model is exposed to missing-support conditions. This is why Sec.~3.4 is critical: artifact-mask injection makes the global loss cover the same defect patterns seen at test time, enabling reliable completion under sparse support.

Input-stage conditioning enforces explicit spatial alignment with low overhead. Under sparse support, this acts as a pixel-level hard constraint at the denoiser input. By contrast, attention-based control tends to treat sparse geometric cues as weaker high-level guidance. This makes the design well-suited for geometry-sensitive extrapolation, where local support patterns are more critical than high-level semantic control.

\subsection{Artifact-Aware Prior Learning for Extrapolation Robustness}
\paragraph{Motivation.}
A core supervision bottleneck in our setting is that we cannot obtain dense, fully reliable RGB targets at truly extrapolated poses during training. Consequently, direct supervised learning on real out-of-trajectory condition--target pairs is intractable. If trained only on self-projected observed-pose pairs, the model mostly sees cleaner geometric conditions than those encountered at extrapolated inference, and thus under-learns recovery under severe support loss. Sec.~3.4 is designed to convert this weakness into a trainable signal: we inject extrapolation-like artifact priors into otherwise aligned training conditions, forcing the generator to learn structure restoration from intentionally degraded geometric evidence.

\paragraph{Design.}
We precompute an artifact mask library $\mathcal{A}=\{a_i\}$ from virtual-pose reprojection, where each $a_i\in\{0,1\}^{H\times W}$ is a validity mask generated by the same reprojection validity test (depth-positive, in-frame, z-buffer-visible). These defects share the same mechanism---support loss after geometric reprojection---and mainly vary in severity and spatial coverage; the resulting mask library therefore provides transferable morphology priors for robust out-of-trajectory synthesis. During training, we sample a Bernoulli variable $b\sim\mathrm{Bernoulli}(p)$; if $b=1$, an artifact mask is drawn uniformly at random from $\mathcal{A}$ and applied to the geometric condition map:
\begin{equation}
\tilde{X}^{v}=\begin{cases}X^{v}\odot a, & b=1,\ a\sim\mathrm{Uniform}(\mathcal{A}) \\ X^{v}, & b=0.\end{cases}
\end{equation}

Artifact injection uses a two-stage sampling scheme. First, a Bernoulli gate with probability $p$ decides whether to perturb the current condition. Second, if activated, one mask is sampled uniformly from the library $\mathcal{A}$. This decouples injection frequency from morphology diversity and avoids over-concentration on a small subset of defect patterns.

Training uses the same mask-and-weighted objective $\mathcal{L}_{\text{train}}$ from Sec.~3.3, with $\tilde{X}^v$ as the conditioning input. Thus, artifact-mask injection acts as condition-space robustness augmentation while leaving the optimization target and timestep sampling unchanged. No extra branch is introduced at inference.

This strategy exposes the model to extrapolation-like defects during optimization and improves robustness under large pose offsets, while keeping inference unchanged.

\subsection{LiDAR-Projected Sparse-Reference (LPSR) Protocol Formulation}
Because dense target-view ground truth is unavailable at extrapolated poses, evaluation must be defined on sparse valid regions. Let $R\in\mathbb{R}^{H\times W\times 3}$ denote the LiDAR-projected sparse reference image and $M\in\{0,1\}^{H\times W}$ its validity mask. We define the valid pixel set
\begin{equation}
\Omega=\{p\mid M(p)=1\}.
\end{equation}
Given a synthesized image $\hat{I}$, all reconstruction metrics are computed only on $\Omega$. The masked error operator is
\begin{equation}
\mathcal{E}_{\Omega}(\hat{I},R)=\frac{1}{|\Omega|}\sum_{p\in\Omega}\left\|\hat{I}(p)-R(p)\right\|_2^2.
\end{equation}
Accordingly, sparse PSNR is defined as
\begin{equation}
\text{S-PSNR}=10\log_{10}\frac{\text{MAX}^2}{\mathcal{E}_{\Omega}(\hat{I},R)},
\end{equation}
and sparse MAE / RMSE follow the same masked-valid-set rule as in Sec.~\ref{sec:experiments}. This formulation ensures that quantitative comparison is restricted to geometrically supported regions and avoids penalizing unverifiable pixels at extrapolated viewpoints.

\subsection{Implementation Details}
\paragraph{Data interface and pair construction.}
Our implementation uses a fixed condition/supervision interface: $X^v$ is a 3-channel reprojected geometric condition map (invalid pixels zero-filled), and $Y^v$ is the RGB supervision image at the same frame--camera identity. Pair matching is deterministic by shared frame and camera indices, independent of variable naming in different code modules.

\paragraph{Geometric validity consistency.}
A single validity operator (depth-positive, in-frame, z-buffer-visible) is reused throughout GAR projection, artifact-mask generation, and sparse-reference evaluation. This shared definition guarantees that condition construction, robustness training, and metric computation are geometrically consistent.

\paragraph{Training objective implementation.}
We follow the single-timestep latent-diffusion objective in Sec.~3.3 with per-sample weighting from the dataloader. Artifact-mask injection only perturbs conditioning inputs and does not alter denoiser architecture or loss form, so training and inference differ only by whether masks are injected.

\section{Experiments}
\label{sec:experiments}

We evaluate Geo-EVS from three complementary perspectives: (i) in-manifold perceptual fidelity, (ii) out-of-manifold extrapolation robustness under sparse geometric support, and (iii) downstream 3D detection utility of synthesized views. Following a unified evaluation protocol, we first define a global baseline pool and then report task-specific subsets based on metric availability and protocol compatibility. All compared methods use identical data splits, preprocessing, and evaluation settings to ensure fairness.

\subsection{Experimental Setup}

\subsubsection{Datasets.}
All experiments are conducted on \textbf{Waymo-mini-1/5} (one-fifth of Waymo v1.3 training data) using fixed train/val/test scene partitions to reduce leakage risk. Geometric pretraining stages (e.g., VGGT adaptation) are performed on scenes disjoint from synthesis evaluation scenes.

\subsubsection{Evaluation Protocol.}
For in-manifold synthesis in this paper, we use FID as the primary perceptual metric (Table~\ref{tab:main_original}), and adopt the LPSR protocol (Sec.~3.5) for extrapolated-view evaluation, where metrics are computed only on LiDAR-projected valid pixels. For downstream evaluation, we follow Waymo's official 3D detection protocol and report LET-mAP$_\mathrm{L}$, LET-mAPH, and L1/mAPH.

\subsubsection{Baselines and Re-implementation.}
We build a global baseline pool spanning reconstruction and generative paradigms: 3DGS~\cite{kerbl20233dgs}, Street Gaussians, EmerNeRF~\cite{emernerf2023}, and FreeVS~\cite{freevs2024}, together with additional generative methods in the observed-view FID comparison (BEVGen as autoregression; Panacea~\cite{wen2024panacea}, MagicDrive, DrivingDiffusion~\cite{li2024drivingdiffusion}, and DriveWM~\cite{wang2024driving} as diffusion). We intentionally use different baseline subsets satisfying protocol requirements per benchmark to avoid selecting baselines that favor our method and to keep comparisons aligned with each evaluation goal. Concretely, Table~\ref{tab:main_original} includes a broader diffusion/autoregressive subset to test whether Geo-EVS remains competitive against methods with strong generative priors under observed-view FID evaluation. In contrast, Table~\ref{tab:extrapolation} and Table~\ref{tab:interpolation} use the same executable core subset (3DGS, EmerNeRF, Street Gaussians, and FreeVS) to provide controlled, protocol-consistent comparisons for sparse-reference extrapolation and NVS interpolation.

All baselines with reproducible and runnable training pipelines are run on our Waymo split with unified preprocessing and evaluation scripts.

\subsubsection{Training and Inference Configuration.}
We train with AdamW (learning rate $1\times10^{-4}$) for a fixed number of iterations. At inference, Geo-EVS uses $T=30$ denoising steps with classifier-free guidance scale $s_{\text{cfg}}=1.5$. Ablations use the same setting. All compared methods share identical input resolution ($512\times512$) and evaluation splits.

\subsection{Perceptual Fidelity Under Observed-View Reconstruction}

We first verify whether extrapolation-oriented training preserves in-manifold visual quality. For each frame, we reconstruct recorded camera views and compare predictions against dense RGB supervision. This experiment serves as a controlled base-quality check before moving to sparse-supervision extrapolation.

Table~\ref{tab:main_original} reports the baseline subset satisfying the protocol requirements for observed-view FID evaluation, covering representative reconstruction, autoregressive, and diffusion methods with available FID. Within this baseline subset, Geo-EVS achieves the best FID, showing that geometry-conditioned diffusion with artifact-aware optimization does not weaken standard in-manifold reconstruction quality.

Compared with strong baselines, Geo-EVS improves markedly over Street Gaussians (20.5), BEVGen (31.27), and prior Stable Diffusion-based methods (best baseline: 14.62). These gains establish a stronger in-manifold starting point for the subsequent out-of-manifold evaluation.

\begin{table}
  \caption{Perceptual comparison on Waymo with method taxonomy. We report each baseline's paradigm and external prior to clarify comparison context; within this baseline subset satisfying the protocol requirements, Geo-EVS achieves the best FID.}
  \label{tab:main_original}
  \centering
  \resizebox{\columnwidth}{!}{%
  \begin{tabular}{lccc}
    \toprule
    Method & Paradigm & Extra Prior & FID $\downarrow$ \\
    \midrule
    Street Gaussians & Reconstruction & None & 20.5 \\
    BEVGen & Autoregression & None & 31.27 \\
    Panacea & Diffusion & Stable Diffusion & 19.63 \\
    MagicDrive & Diffusion & Stable Diffusion & 18.45 \\
    DrivingDiffusion & Diffusion & Stable Diffusion & 17.81 \\
    DriveWM & Diffusion & Stable Diffusion & 14.62 \\
    \rowcolor{gray!12}
    Geo-EVS (Ours) & Diffusion & Stable Diffusion & \textbf{3.9} \\
    \bottomrule
  \end{tabular}%
  }
\end{table}

All methods listed in Table~\ref{tab:main_original} provide FID values, so ranking statements are made directly on complete entries.

\subsection{Robustness Under Out-of-Manifold Extrapolation}

We first evaluate the key target regime: out-of-manifold extrapolation with limited geometric support. Table~\ref{tab:extrapolation} reports the extrapolation-compatible baseline subset under a unified sparse-reference protocol. Concretely, we use the front ($0^\circ$) and front-side ($45^\circ$) cameras to construct an intermediate view at $22.5^\circ$, and then apply 1~m lateral shifts in the world coordinate system to both left and right directions to obtain extrapolated targets. This setting explicitly departs from pure interpolation and directly measures robustness under pose extrapolation.

\textbf{Protocol and metrics.} Because dense RGB supervision is unavailable after the lateral translation, we follow the LPSR protocol and evaluate only on LiDAR-projected valid pixels. We report sparse-PSNR and sparse-SSIM, which measure reconstruction fidelity on geometrically verifiable regions and avoid penalizing unverifiable pixels in unsupported areas.
\begin{table*}[t]
  \caption{Extrapolation robustness on Waymo. \textbf{Top:} Overall method comparison. \textbf{Bottom:} Geo-EVS detailed analysis by pose offset and geometric sparsity, showing consistent performance across extrapolation conditions.}
  \label{tab:extrapolation}
  \centering

  \begin{tabular*}{0.90\textwidth}{@{\extracolsep{\fill}}lcc}
    \toprule
    \textbf{Method} & \textbf{sparse-PSNR} $\uparrow$ & \textbf{sparse-SSIM} $\uparrow$ \\
    \midrule
    3DGS & 19.840 & 0.824 \\
    EmerNeRF & 21.850 & 0.876 \\
    Street Gaussians & 23.000 & 0.932 \\
    FreeVS & 22.920 & 0.895 \\
    \midrule
    \rowcolor{gray!12}
    Geo-EVS (Ours) & \textbf{23.650} & \textbf{0.941} \\
    \bottomrule
  \end{tabular*}

  \vspace{0.25cm}

  \begin{tabular*}{0.90\textwidth}{@{\extracolsep{\fill}}lccccc|ccc}
    \toprule
    \multicolumn{9}{c}{\textbf{Geo-EVS Detailed Analysis (Ours)}} \\
    \cmidrule{1-9}
    & \multicolumn{5}{c|}{\textbf{By Pose Offset ($^\circ$)}} & \multicolumn{3}{c}{\textbf{By Sparsity}} \\
    \cmidrule(lr){2-6} \cmidrule(lr){7-9}
    \textbf{Metric} & [0-5] & [5-10] & [10-15] & [15-20] & [20-30] & [0-0.02] & [0.02-0.05] & [0.05-0.1] \\
    \midrule
    sparse-PSNR $\uparrow$ & 24.17 & 23.80 & 23.57 & 23.42 & 23.30 & 24.39 & 20.73 & \textbf{26.37} \\
    sparse-SSIM $\uparrow$ & 0.944 & 0.942 & 0.940 & 0.939 & 0.938 & \textbf{0.984} & 0.926 & 0.955 \\
    \bottomrule
  \end{tabular*}
\end{table*}

Within the baseline subset satisfying the protocol requirements in Table~\ref{tab:extrapolation}, Geo-EVS achieves the best sparse-PSNR and sparse-SSIM, outperforming both reconstruction-based and diffusion-based baselines under the same sparse-reference protocol. In particular, compared with the strongest reconstruction baseline (Street Gaussians), our method yields consistent gains on both metrics, indicating that geometry-conditioned diffusion with artifact-aware training better handles support discontinuities introduced by lateral extrapolation.

The detailed analysis in the lower part of Table~\ref{tab:extrapolation} further validates robustness trends. As the pose offset increases, performance degrades smoothly rather than collapsing, suggesting stronger tolerance to wider viewpoint gaps. Across sparsity bins, the model maintains stable quality under reduced valid support, consistent with our objective of exposing training to reprojection-induced defect patterns.

\begin{figure*}[t]
  \centering
  \safeimg{0.87\textwidth}{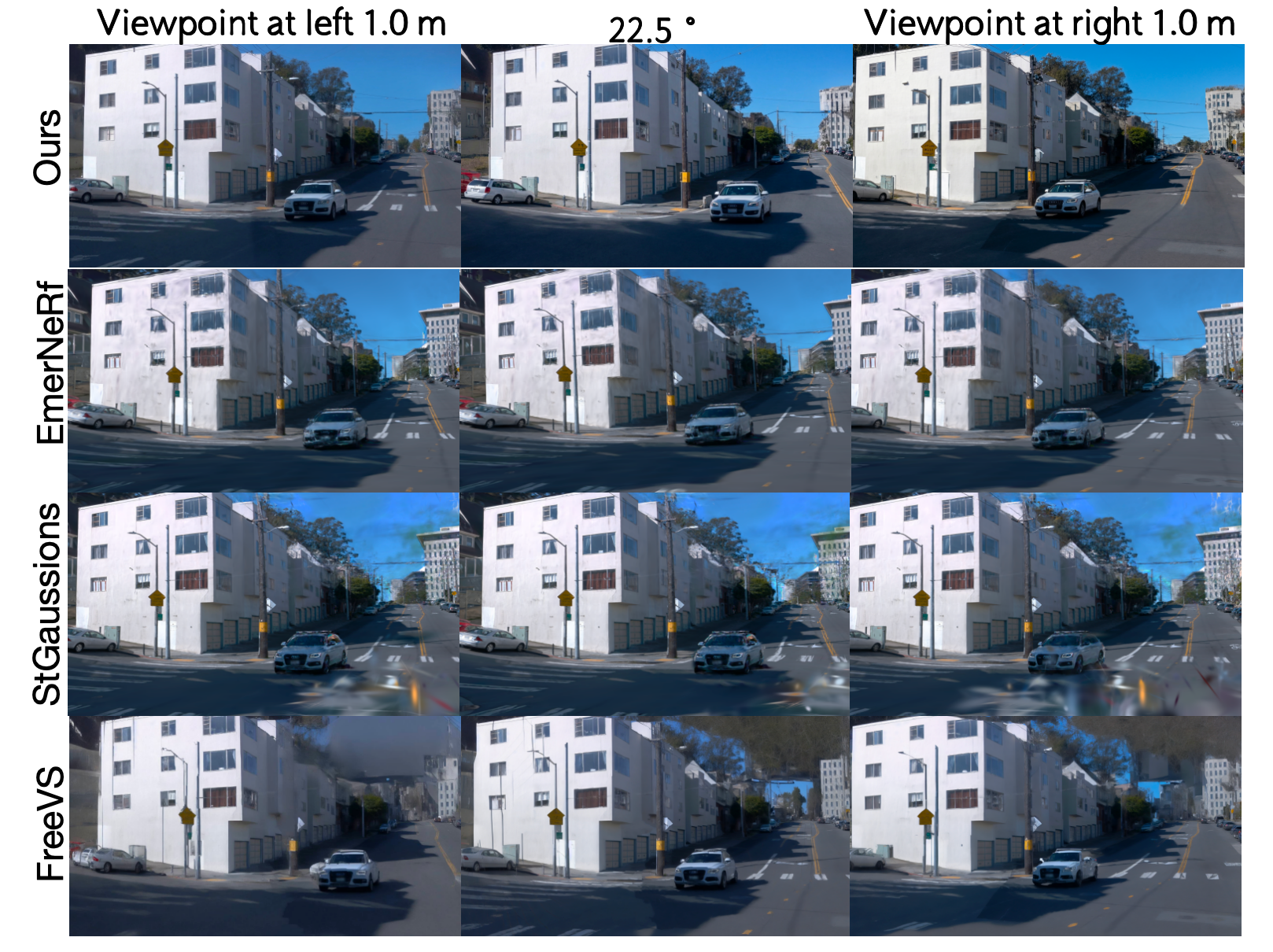}
  \caption{Qualitative extrapolation comparison aligned with Table~\ref{tab:extrapolation}. We first construct an intermediate target at $22.5^{\circ}$ from the front ($0^{\circ}$) and front-side ($45^{\circ}$) views, then apply 1~m left/right lateral translations in the world frame to form out-of-manifold targets. Geo-EVS yields more coherent structures and cleaner local details than competing methods under this setting.}
  \Description{A grid of street-view images comparing the extrapolation synthesis results of five methods: 3DGS, EmerNeRF, Street Gaussians, FreeVS, and the proposed Geo-EVS. The rows show different virtual views generated by applying 1-meter lateral shifts (left and right) from an intermediate 22.5-degree angle. Columns represent the different methods. The comparison shows that while baseline methods like 3DGS and EmerNeRF produce blurry or distorted structures, and FreeVS hallucinates incorrect details, the proposed Geo-EVS maintains clear edges, coherent geometry, and cleaner local details on elements like lane markings and vehicle shapes under large viewpoint shifts.}
  \label{fig:extrapolation_vis}
\end{figure*}

Overall, these results verify that Geo-EVS improves not only average extrapolation fidelity but also stability in harder geometric conditions, which is essential for reliable virtual-view deployment. Qualitative comparisons in Fig.~\ref{fig:extrapolation_vis} are consistent with the quantitative gains in Table~\ref{tab:extrapolation}.

\subsection{In-Manifold View Interpolation}

\begin{table*}
  \caption{Downstream 3D detection with BEVFormer on Waymo-mini-1/5. We compare the default 5-view setting and 45-view setting with 40 generated views. All metrics are evaluated on the Waymo validation split.}
  \label{tab:detection_single}
  \centering
  \begin{tabular}{lccc}
    \toprule
    Setting & LET-mAP$_\mathrm{L}$ $\uparrow$ & LET-mAPH $\uparrow$ & L1/mAPH (Car) $\uparrow$ \\
    \midrule
    Original 5 Views & 34.9 & 46.3 & 25.5 \\
    5 Native + 40 Generated Views (Ours) & \textbf{35.9} & \textbf{47.1} & \textbf{26.8} \\
    Gain & +1.0 & +0.8 & +1.3 \\
    \bottomrule
  \end{tabular}
\end{table*}

\begin{table*}
  \caption{Core ablation study on extrapolative synthesis (evaluated on view extrapolation split). The three-variant design isolates artifact-prior quality under fixed concat conditioning.}
  \label{tab:ablation_core}
  \centering
  \begin{tabular}{lcccccc}
    \toprule
    Variant & Artifact Lib. & Mask Type & $p$ & Conditioning & S-PSNR $\uparrow$ & Contribution \\
    \midrule
    V1: Baseline (no artifacts) & \xmark & -- & 0.0 & Concat & 21.73 & baseline \\
    V2: + Random masks & \xmark & Random Box & 0.4 & Concat & 22.34 & +0.61 dB \\
    \rowcolor{gray!12}
    V3: Full model (Reproj masks) & \cmark & Reproj Mask & 0.4 & Concat & \textbf{23.65} & \textbf{+1.92 dB} \\
    \bottomrule
  \end{tabular}
\end{table*}

We next examine whether optimizing for extrapolation compromises standard in-manifold synthesis. Following Table~\ref{tab:interpolation}, we use all front-side camera views as source inputs and synthesize the front camera view as target.

\begin{table}
  \caption{Comparison with NVS counterparts on novel camera synthesis. For all NVS methods, we use all front-side camera views as source views to synthesize the front camera views.}
  \label{tab:interpolation}
  \centering
  \begin{tabular}{lccc}
    \toprule
    Methods & SSIM $\uparrow$ & PSNR $\uparrow$ & LPIPS $\downarrow$ \\
    \midrule
    3DGS & 0.624 & 18.06 & 0.335 \\
    Street Gaussians & 0.741 & 21.72 & 0.221 \\
    EmerNeRF & 0.678 & 19.51 & 0.287 \\
    FreeVS & \textbf{0.768} & 22.85 & 0.179 \\
    \rowcolor{gray!12}
    Geo-EVS (Ours) & 0.736 & \textbf{23.42} & \textbf{0.154} \\
    \bottomrule
  \end{tabular}
\end{table}

As shown in Table~\ref{tab:interpolation}, Geo-EVS remains highly competitive on in-manifold synthesis and delivers the strongest PSNR/LPIPS trade-off among compared methods. Fig.~\ref{fig:interpolation_vis} provides visual comparisons aligned with this protocol. These results indicate that the gains observed in extrapolation are not obtained by sacrificing interpolation performance; instead, the proposed design preserves conventional NVS quality while improving out-of-manifold robustness.

\subsection{Qualitative Evidence}

\begin{figure*}[t]
  \centering
  \safeimg{0.87\textwidth}{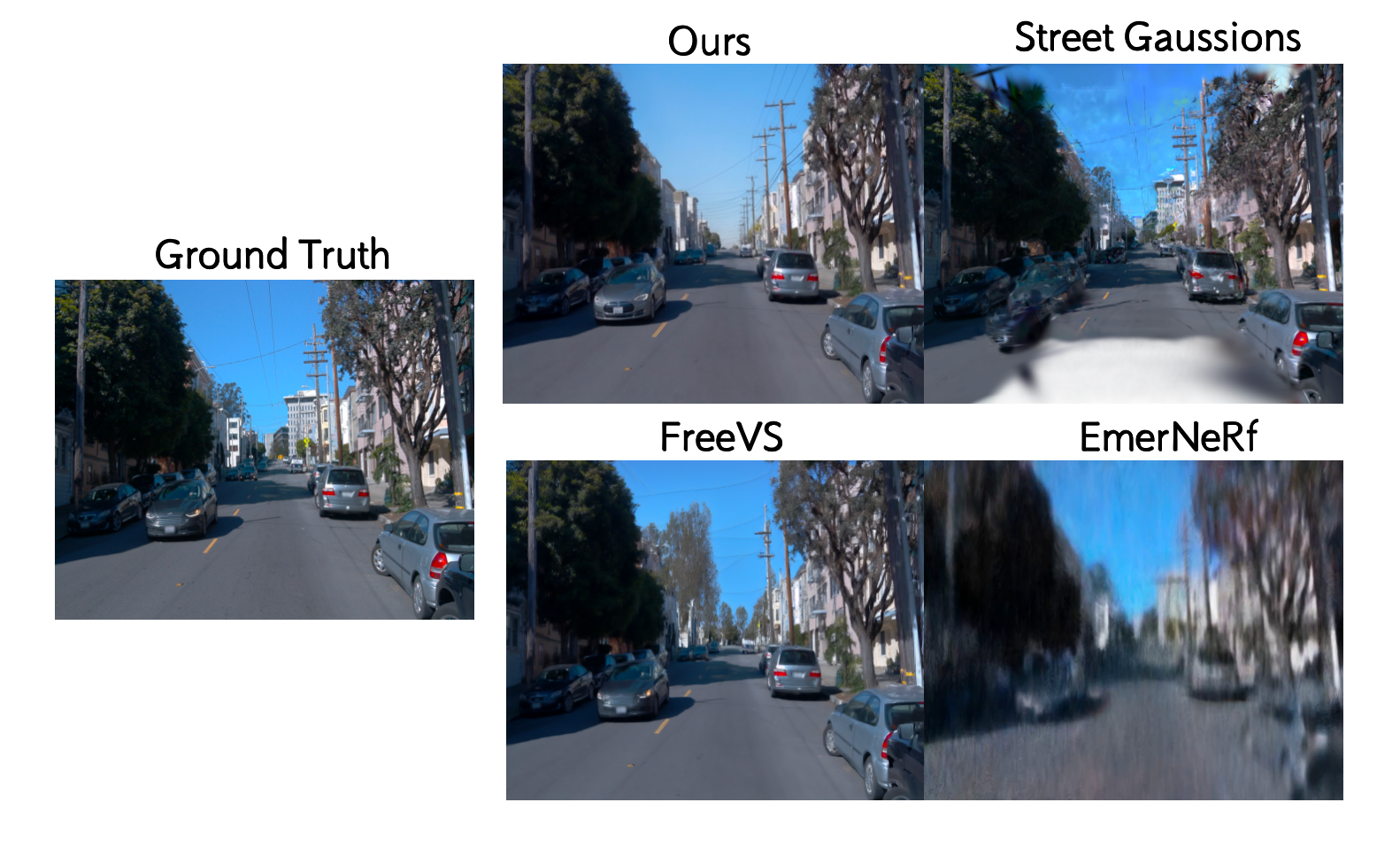}
  \caption{Qualitative interpolation comparison aligned with Table~\ref{tab:interpolation}. Following the same protocol, all methods take only front-side camera views as input and synthesize the front camera view. Geo-EVS preserves sharper structures and more consistent textures while remaining competitive with strong NVS baselines.}
  \Description{A grid of street-view images comparing the interpolation synthesis results of the same five methods: 3DGS, EmerNeRF, Street Gaussians, FreeVS, and Geo-EVS. The task is to synthesize the front camera view using only front-side camera inputs. The columns compare the outputs of each method alongside the Ground Truth image. The visual results indicate that Geo-EVS produces sharper structural boundaries and more consistent textures compared to 3DGS and Street Gaussians, and is highly competitive with EmerNeRF and FreeVS in maintaining high-fidelity appearance in overlapping viewpoint regions.}
  \label{fig:interpolation_vis}
\end{figure*}

To assess whether the quantitative gains correspond to meaningful structural improvements, we compare extrapolated view generations across methods in Fig.~\ref{fig:extrapolation_vis} and in-manifold interpolation quality in Fig.~\ref{fig:interpolation_vis}.

Reconstruction-based methods (3DGS, Street Gaussians) show topology breaks and stretched structures in unsupported regions. EmerNeRF produces floating artifacts and blur. FreeVS, although diffusion-based, still hallucinates content inconsistent with sparse geometric support under large extrapolation. In contrast, Geo-EVS better preserves geometric layout while recovering plausible texture. Edges are more aligned with projected structure, valid-support regions remain sharper, and missing regions are completed with fewer structural conflicts. This visual evidence is consistent with the sparse-reference gains in Table~\ref{tab:extrapolation}.

\subsection{Downstream Utility for 3D Detection}

Beyond image-level quality, we further evaluate whether synthesized views improve downstream perception. Table~\ref{tab:detection_single} reports a controlled BEVFormer study comparing the default 5-view setting against a 45-view setting with 40 generated views.

\textbf{Setup.} We train BEVFormer on Waymo-mini-1/5 and evaluate on the Waymo validation split. All settings share the same detector architecture, optimization schedule, and preprocessing pipeline; only camera input configuration changes (5 native views vs. 5 native + 40 generated views). We report LET-mAP$_\mathrm{L}$, LET-mAPH, and L1/mAPH (Car). All results use PNG-format data.

The 45-view setting consistently improves all metrics: +1.0 LET-mAP$_\mathrm{L}$, +0.8 LET-mAPH, and +1.3 L1/mAPH. The concurrent gains in localization and heading quality indicate that generated views provide complementary geometric cues rather than redundant appearance information.

\subsection{Ablation Causality}

To identify which design choice is responsible for extrapolation gains, we perform targeted ablations tied to our core claim that artifact-aware robustness training is a key factor for extrapolation quality.

Relative to V1, reprojection-derived masks (V3) improve S-PSNR by +1.92 dB, substantially larger than random-mask augmentation (V2, +0.61 dB). This indicates that morphology-matched defect priors, not generic corruption alone, are responsible for most robustness gains.

\subsection{Failure Boundaries and Limitations}

Geo-EVS still has three main failure modes: (i) \emph{dynamic-object inconsistency} when moving objects appear at conflicting positions across source views; (ii) \emph{severe hallucination} under extremely sparse support (e.g., $<5\%$ valid pixels), where priors dominate geometry; and (iii) \emph{texture bleeding} near occlusion boundaries when point-cloud detail is insufficient. These boundaries clarify where current gains may degrade and motivate future extensions in temporal consistency and multi-frame fusion.

\section{Conclusion}
This paper studies extrapolative view synthesis for autonomous driving under sparse supervision and introduces \textbf{Geo-EVS}, a geometry-conditioned generation framework for this setting. The core insight is to treat virtual-pose reprojection defects as a structured projection-domain shift and to train the generator explicitly on this shift rather than relying on clean observed-pose conditions alone. Concretely, GAR preserves a unified reprojection interface between training and inference, and AGLD with artifact-aware mask injection improves robustness to missing geometric support.

Experiments on Waymo validate this design from three perspectives: in-manifold perceptual quality, out-of-manifold extrapolation robustness under sparse-reference evaluation, and downstream utility for 3D detection. Across these evaluations, Geo-EVS achieves consistent gains over representative reconstruction and diffusion baselines, with the largest advantages appearing in high-offset and low-support regimes where extrapolation is most difficult.

A current limitation is that performance still degrades under extreme sparsity and highly dynamic scenes, where temporal inconsistency and structural hallucination can remain. Extending the framework with stronger temporal modeling and multi-frame geometric fusion is an important next step toward more reliable long-horizon deployment.

\section*{Acknowledgments}
This work was supported by Chongqing Changan Automobile Co., Ltd., Project ``Research on Controllable Long-Horizon Video Generation for Autonomous Driving Scenarios Based on World Models'' (Grant No. 20252002218).
\bibliographystyle{unsrt}
\bibliography{lyt}











\end{document}